\documentclass{article} % For LaTeX2e
\usepackage{iclr2027_conference,times}

\usepackage{amsmath,amsfonts,bm}

\def\eqref#1{equation~\ref{#1}}
\def\1{\bm{1}}

\DeclareMathAlphabet{\mathsfit}{\encodingdefault}{\sfdefault}{m}{sl}
\SetMathAlphabet{\mathsfit}{bold}{\encodingdefault}{\sfdefault}{bx}{n}

\usepackage{hyperref}
\usepackage{url}

\usepackage{booktabs}
\usepackage{multirow}
\usepackage{graphicx}
\usepackage{adjustbox}

\title{\textbf{FATE-5}: \textbf{F}aithful \textbf{A}lignment of \textbf{T}rait \textbf{E}xpression for Continuous Big Five Personality Control}

\author{Gunhee Cho, Hongyeob Kim \& Yun-Gyung Cheong \\
Department of Artificial Intelligence\\
Sungkyunkwan University\\
Suwon, South Korea \\
\texttt{\{skate4333,bongyeob2,aimecca\}@skku.edu}\\
}

\iclrfinalcopy % Uncomment for camera-ready version, but NOT for submission.
\begin{document}

\maketitle

\lhead{} % arxiv 올리기 위해서 설정

\begin{abstract}
Controlling personality in large language models is important for building
conversational agents that can be customized to different users and interaction
settings. However, existing personality control methods often face a tradeoff
between strong trait expression and natural dialogue generation. We introduce
\textbf{FATE-5}, Faithful Alignment of Trait Expression for the Big Five, a
two-stage framework that separates personality encoding from naturalness
alignment. FATE-5 first learns continuous soft prompt representations of Big Five
profiles using dialogue supervision and an auxiliary survey objective. It then
improves dialogue naturalness through preference alignment based on both dialogue
quality and personality agreement. Across three language model backbones,
FATE-5 maintains strong trait expression while achieving substantially higher
dialogue naturalness than other trait-strong methods. On held-out questionnaires,
it achieves monotonicity correlations of 0.97--0.99 and recovery RMSE of
8.9--12.4, substantially improving over the evaluated baselines. These results
demonstrate effective continuous personality control while maintaining natural
conversational behavior.
\end{abstract}

\section{Introduction}

As large language models increasingly serve as conversational agents, controlling
their personality is increasingly important for adapting them to different users and
interaction settings~\citep{psychadapter}. 
%Personality control turns behavioral characteristics into explicit design variables and enables fine grained customization of how an agent communicates~\citep{psychadapter}. 
Such adaptation can be particularly useful in
interactive applications, where personality can influence dialogue behavior,
task success, and user satisfaction~\citep{guo2025personality}. 
The Big Five
provides a natural framework for this purpose by organizing personality into five
broad and continuous dimensions of Openness, Conscientiousness, Extraversion,
Agreeableness, and Neuroticism~\citep{big5}. Its continuous structure provides an interpretable way to specify the trait configuration and intensity an agent should express.

Recent work has explored several approaches for inducing Big Five personality in
LLMs. Prompt-based methods use natural language descriptions or explicit
personality profiles to elicit desired traits~\citep{personallm}.
Training-based approaches instead learn personality conditioned model
representations or adaptations~\citep{dpg,big5chat}, while activation-based
methods directly manipulate internal directions associated with personality
traits~\citep{steering-vector}. 
%These studies demonstrate that LLM behavior can be shifted toward desired personality profiles, but existing methods exhibit different limitations. 
Prompting tends to preserve natural conversation while providing relatively weak trait expression. Training based methods offer more expressive conditioning mechanisms, although accurately matching the intended continuous trait intensity remains challenging. Activation steering provides flexible and strong control without retraining the underlying model. However, our interactive evaluation shows that strong intervention can substantially degrade dialogue naturalness. These limitations reveal a tradeoff between personality control strength and dialogue naturalness that existing approaches do not fully resolve.
%Prompting tends to preserve natural conversation but provides relatively weak trait expression. Training based methods provide more expressive conditioning mechanisms, but accurately preserving the intended continuous trait intensity remains challenging. Activation steering provides flexible and strong control without retraining the underlying model, but our interactive evaluation shows that strong intervention can substantially degrade dialogue naturalness. These limitations reveal an important tradeoff between personality control strength and conversational quality that existing approaches do not fully resolve.

To address this tradeoff, we propose \textbf{FATE-5}, Faithful Alignment of Trait
Expression for the Big Five, a framework for faithful and natural continuous
personality control. Given a target Big Five profile
$\bm{b}\in[0,100]^5$, FATE-5 first maps the profile into continuous soft prompt
embeddings using a lightweight personality encoder while keeping the pretrained
language model frozen, following the general principle of parameter efficient
prompt tuning~\citep{prompt-tuning}. The encoder is trained using two
complementary objectives. Dialogue supervision teaches how the target personality
should be expressed in conversation, while an auxiliary survey objective
strengthens the correspondence between continuous target values and their learned
representation. This objective is motivated by psychometric studies showing that
standardized personality instruments can provide structured measurements of
personality expressed by LLMs~\citep{serapiogarcia}. We find,
however, that stronger trait control can reduce dialogue naturalness. FATE-5
therefore separates personality encoding from naturalness alignment. After
learning the personality representation, we freeze the encoder and attach a LoRA
adapter to the backbone~\citep{lora}. We construct preference pairs that
jointly consider dialogue naturalness and agreement with the target personality,
and optimize the adapter using Direct Preference
Optimization~\citep{rafailov2023dpo}. This two stage design improves
conversational quality while preserving strong personality control. We evaluate
FATE-5 through complementary interactive dialogue and psychometric protocols,
measuring both perceptible trait expression and recovery of the intended
continuous personality profile.

Our contributions are as follows.

\begin{itemize}
%\item We study continuous Big Five personality control from the joint perspective of trait expression and dialogue quality. 
\item We characterize continuous Big Five personality control along two dimensions: trait expression and dialogue quality. %Through a common interactive evaluation, 
Using a multi-turn dialogue evaluation, we identify a tradeoff between strong personality induction and natural conversation in existing approaches.

\item We introduce \textbf{FATE-5}, a two-stage framework that separates personality encoding from naturalness alignment. The first stage learns soft prompt representations of continuous Big Five profiles using dialogue and survey supervision. The second stage improves conversational naturalness through preference alignment based on both dialogue quality and personality agreement. %We introduce \textbf{FATE-5}, a two-stage framework that separates personality encoding from naturalness alignment.
%The first stage maps continuous Big Five profiles to soft prompt representations using a lightweight personality encoder while keeping the pretrained backbone frozen. Dialogue supervision and an auxiliary survey objective improve how faithfully the learned representation encodes the target personality. 
%We introduce \textbf{FATE-5}, a lightweight framework that maps continuous Big Five profiles to soft prompt representations while keeping the pretrained backbone frozen. Dialogue supervision and an auxiliary survey objective strengthen the correspondence between the target profile and the induced personality.

%\item The second stage uses preference alignment based on both dialogue naturalness and personality agreement, improving naturalness while preserving the target personality.%We introduce a preference alignment stage that improves conversational naturalness while preserving the intended personality by jointly considering dialogue quality and trait agreement. %through a preference signal that jointly considers dialogue quality and trait agreement.

\item We evaluate FATE-5 using dialogue and psychometric protocols across multiple backbones and questionnaire formats, with ablations of its key components, assessing both personality expression and control fidelity. %We evaluate FATE-5 using complementary dialogue and psychometric protocols across multiple backbones, questionnaire formats, continuous target values, and component ablations, providing a broader assessment of personality expression and control fidelity.

\end{itemize}

\section{Related Work}

\subsection{Personality Control in Large Language Models}

A growing body of work studies how personality can be imposed on or elicited
from language models. The simplest approach specifies personality through
natural-language instructions. Big5-Scaler encodes numerical Big Five scores
directly into prompts and demonstrates that personality intensity can be
controlled without parameter updates~\citep{cho2025scaling}. PersonaLLM
similarly studies whether prompted LLM personas can express assigned Big Five
characteristics across personality questionnaires and open-ended
generation~\citep{personallm}. Prompt based approaches are inexpensive and model agnostic, but they often provide weaker trait expression than more direct intervention methods, limiting their ability to induce clear and consistent personality differences.

Other approaches modify model representations or parameters more directly. DPG
uses a hypernetwork to generate personality-conditioned low-rank model
adaptations~\citep{dpg}. PsychAdapter learns layer-specific
transformations from continuous psychological attributes and conditions
transformer layers on these representations~\citep{psychadapter}.
Activation-steering approaches instead identify directions associated with
personality traits in a pretrained model and intervene directly in its hidden
states. PERSONA extracts trait-specific activation directions and supports
continuous and compositional personality control through operations in activation
space~\citep{steering-vector}. More recent activation-space work also studies how
layer selection affects the stability of Big Five steering
~\citep{bhandari2026}. These approaches provide strong personality control, but their conversational quality is less clear. In our experiments, we find that stronger trait control can reduce dialogue naturalness. FATE-5 is designed to address this tradeoff by maintaining strong personality expression while improving natural dialogue quality.

% Personality-conditioned generation also depends on data connecting language to
% personality. PANDORA provides Reddit text associated with author-level
% personality and demographic information~\citep{pandora}, while
% BIG5-CHAT provides human-grounded dialogue constructed to express controlled Big
% Five characteristics~\citep{big5chat}. We use the two sources for
% complementary purposes. BIG5-CHAT supplies strongly expressed personality
% signals, whereas PANDORA provides naturally occurring language paired with
% continuous trait scores. Unlike approaches that learn personality expression
% from dialogue supervision alone, our training additionally introduces
% questionnaire supervision to strengthen the fidelity of the resulting
% personality representation.

\subsection{Psychometric Evaluation of LLM Personality}

Standardized personality inventories have long been used to measure the Big Five
traits. The Big Five Inventory provides a widely used measure of Openness,
Conscientiousness, Extraversion, Agreeableness, and
Neuroticism~\citep{bfi}. The NEO Five Factor Inventory offers another
established measure of the five broad dimensions~\citep{neo-ffi}, while the
IPIP NEO 120 provides a public domain inventory covering both broad traits and
their lower level facets~\citep{ipip-neo-120}. Shorter instruments such as the
Ten Item Personality Inventory enable more compact assessment
~\citep{tipi}, and the BFI-2 extends the original BFI with a
hierarchical structure that includes additional facet level information
~\citep{bfi-2}.

These psychometric instruments have also been increasingly used to measure
personality in LLMs. PersonaLLM evaluates assigned Big Five personas using the
BFI together with open ended writing tasks~\citep{personallm}.
\citet{serapiogarcia} develop a broader psychometric
framework for administering personality inventories to LLMs and evaluating the
reliability and construct validity of their responses. TRAIT further introduces
a personality benchmark for LLMs using psychometrically grounded questions
derived from established personality constructs~\citep{trait}. Other work
compares Likert and forced choice formats and shows that the response format can
substantially affect estimated LLM personality
~\citep{li-etal-2025-decoding-llm}. Together, these studies establish
standardized questionnaires as an important tool for analyzing personality in
language models.

At the same time, psychometric evaluation of LLMs remains sensitive to how the
assessment is conducted. Personality measurements can vary with prompt wording
and response format, and widely available questionnaire items may also appear in
model pretraining data~\citep{serapiogarcia, han-etal-2026-quantifying}. These findings
motivate evaluating personality with multiple instruments and question formats
rather than relying on a single questionnaire setting. In our experiments, we
therefore use separate inventories for training and evaluation and complement
questionnaire based measurement with dialogue evaluation.

% \subsection{Preference Alignment}

% Preference based alignment has been widely used to adapt language model behavior to
% human preferences. Earlier approaches such as RLHF typically train a reward model
% from preference data and then optimize the language model using reinforcement
% learning~\citep{rlhf}. Direct Preference Optimization simplifies
% this process by learning directly from preferred and dispreferred responses
% without fitting an explicit reward model~\citep{rafailov2023dpo}. In our
% framework, DPO is applied after personality encoding to improve dialogue
% naturalness while preserving the induced personality.

% ======================================================================
%  Method  (requires: amsmath, amssymb, bm)
% ======================================================================

\begin{figure}[t!]
    \centering
    \includegraphics[width=1\columnwidth]{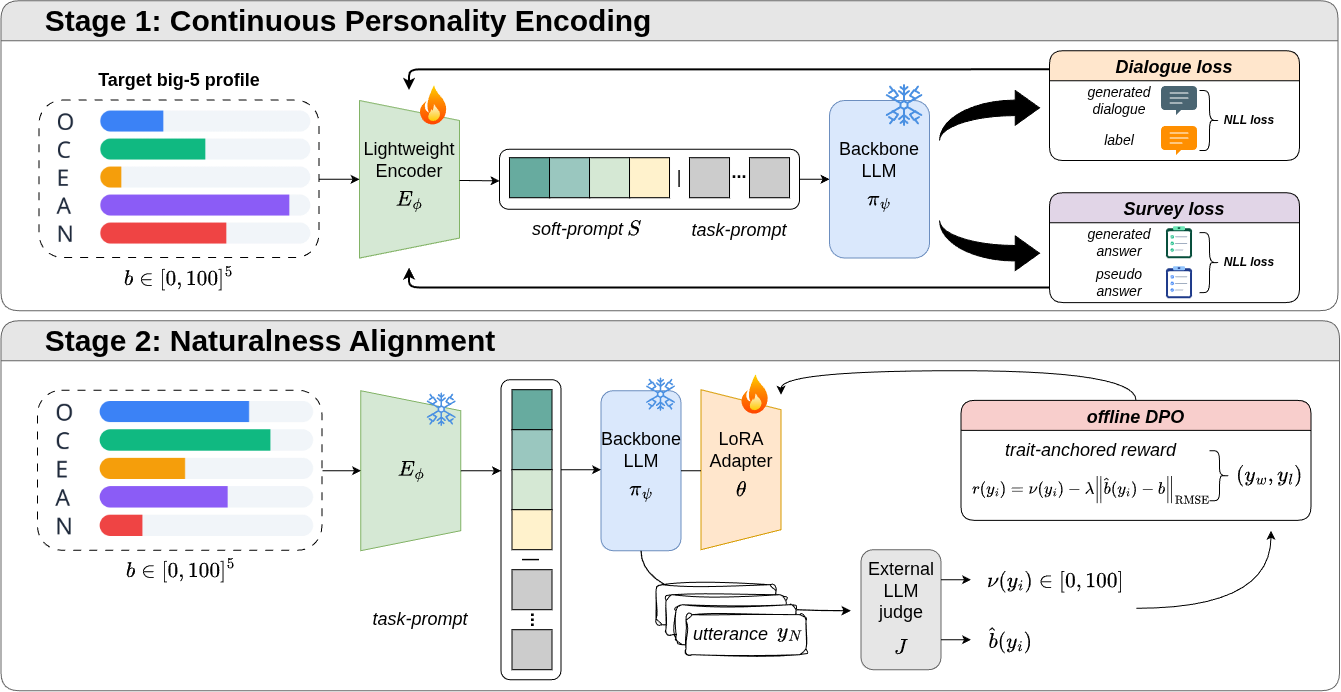}
\caption{Overview of FATE-5. Stage 1 learns a continuous personality representation from a target Big Five profile using a lightweight encoder while keeping the backbone LLM frozen. The encoder is trained jointly with dialogue loss and survey loss to strengthen trait expression. Stage 2 freezes the encoder and backbone, attaches a trainable LoRA adapter, and performs offline DPO using preference pairs scored by an external LLM judge. This stage improves dialogue naturalness while preserving the personality representation learned in Stage 1.}
    \label{fig:fate5}
\end{figure}

\section{Method}
\label{sec:method}

\subsection{Problem Setup}
\label{sec:setup}
We consider inducing a target personality in an instruction-tuned language
model along the Big Five (OCEAN) dimensions. A target profile is a vector
$\bm{b}=(b_\mathrm{O},b_\mathrm{C},b_\mathrm{E},b_\mathrm{A},b_\mathrm{N})\in[0,100]^5$,
one score per trait. Given $\bm{b}$ and a dialogue context $c$, a personality conditioning method
defines a policy $\pi(\cdot\mid \bm{b},c)$ that generates an utterance.
\textbf{Our goal is that the generated dialogue faithfully and naturally
expresses $\bm{b}$}. The target trait levels should be perceptible in the
conversation while it remains natural and engaging.

% Faithful expression is the central difficulty. Existing methods such as prompting, steering vectors, and additive adapters instantiate $\pi$ by perturbing the input text or hidden activations. These methods can shift the speaking style, but the target trait is often encoded weakly or inconsistently. As a result, the personality perceived by a reader can drift from $\bm{b}$. Our method directly addresses this fidelity problem using a personality encoder whose trait encoding is strengthened by an auxiliary survey objective (\S\ref{sec:objective}). We then apply a preference-alignment stage to recover the naturalness degraded by the survey objective (\S\ref{sec:dpo}).

\subsection{Personality Encoder}
\label{sec:encoder}
We introduce a lightweight encoder
$E_\phi:[0,100]^5\!\rightarrow\!\mathbb{R}^{T\times d}$ that maps a target profile
to a sequence of $T$ continuous \emph{soft-prompt} embeddings
\begin{equation}
  \bm{S} \;=\; E_\phi(\bm{b}) \;=\; \big(\bm{s}_1,\dots,\bm{s}_T\big),
  \qquad \bm{s}_t\in\mathbb{R}^{d},
\end{equation}
where $d$ is the model hidden size. The encoder is a two-layer MLP
with GELU activations. Let $\pi_\psi$ be a pretrained, \emph{frozen} language model with token embedding
matrix $\bm{W}_e$. Given a tokenized input $x$ with embeddings
$\bm{E}(x)=\bm{W}_e[x]$, we prepend the soft prompt and run the frozen model on
the concatenated sequence,
\begin{equation}
  \pi(\cdot\mid \bm{b},x) \;=\;
  \pi_\psi\!\big([\,\bm{S};\,\bm{E}(x)\,]\big).
  \label{eq:prepend}
\end{equation}
Only $\phi$ is trainable; the backbone $\psi$ never receives gradients. This
keeps the personality signal confined to the input representation and makes the
method parameter-efficient and backbone-agnostic.

\subsection{Training Objective: Dialogue and Survey}
\label{sec:objective}
We train $E_\phi$ with two complementary losses.

\paragraph{Dialogue loss.} On a corpus of $(\bm{b}, x, y)$ triples, where $\bm{b}$ denotes the personality profile, $x$ the prompt, and $y$ the reference response, we minimize the negative log-likelihood of the response under the conditioned policy,
\begin{equation}
  \mathcal{L}_\mathrm{dial}(\phi)
   = -\,\mathbb{E}_{(\bm{b},x,y)}
     \sum_{k} \log \pi\!\big(y_k \mid \bm{b},x,y_{<k}\big).
\end{equation}
This teaches the soft prompt to shape generation toward $\bm{b}$.

\paragraph{Survey loss.} The dialogue loss shapes style, but on its own it encodes the target only weakly, causing the trait perceived by a reader to drift from $\bm{b}$. To improve how faithfully the soft prompt encodes the target, we add an auxiliary survey loss that trains the conditioned model to report $\bm{b}$ under standardized Big Five questionnaires. For a standardized questionnaire with items ${q_j}$ and an answer scale, let $\rho(\cdot)$ denote the questionnaire's scoring rubric that maps item responses to trait scores. We derive pseudo-answers $\tilde{a}_j$ from the target $\bm{b}$ so that they are consistent with the target profile, and train the conditioned model to produce them. The construction of $\tilde{a}_j$ is detailed in Appendix~\ref{app:pseudo}.
\begin{equation}
  \mathcal{L}_\mathrm{surv}(\phi)
   = \mathbb{E}_{\bm{b}}\;\sum_{j}
     \ell\!\big(\pi(\cdot\mid \bm{b},q_j),\,\tilde{a}_j\big(\bm{b})),
  \label{eq:surv}
\end{equation}
where $\ell$ is a token-level loss on the rated answer. During training, we draw items from a set of training questionnaires. This auxiliary objective serves as a means to strengthen personality expression. The survey loss directly encourages the soft-prompt representation to preserve the target trait values.

\paragraph{Full objective.} The encoder is trained end-to-end with
\begin{equation}
  \min_{\phi}\;\;
  \lambda_\mathrm{dial}\,\mathcal{L}_\mathrm{dial}(\phi)
  \;+\;
  \lambda_\mathrm{surv}\,\mathcal{L}_\mathrm{surv}(\phi),
  \qquad \psi\ \text{frozen}.
  \label{eq:full}
\end{equation}

\subsection{Trait-Preserving Naturalness Alignment via Offline DPO}
\label{sec:dpo}
The encoder produces strong, measurable trait control, but prepending a soft
prompt to a frozen backbone incurs a cost in fluency: conditioned generations
are less natural than the unconditioned model. To recover naturalness
\emph{without} sacrificing the trait, we add a stage of offline preference
alignment on top of the frozen encoder.

We attach a low-rank adapter (LoRA) $\theta$  to the frozen backbone $\pi_\psi$ (the pretrained LLM of §3.2, whose own weights $\psi$ stay fixed), giving a policy
$\pi_\theta(\cdot\mid\bm{b},x)$ that still uses the frozen soft prompt
$\bm{S}=E_\phi(\bm{b})$ but now has trainable weights $\theta$. The reference
policy $\pi_\mathrm{ref}$ is the same model with the adapter disabled.

\paragraph{Preference pairs.} For each sampled profile $\bm{b}$ and topic, we
draw $N$ candidate utterances $\{y_i\}$ from the encoder policy. An external LLM
judge $J$ rates each candidate in a single call, returning a naturalness score
$\nu(y_i)\in[0,100]$ and an estimated profile $\hat{\bm{b}}(y_i)$. We score each
candidate with a trait-anchored reward
\begin{equation}
  r(y_i)\;=\;\nu(y_i)\;-\;\lambda\,\big\|\hat{\bm{b}}(y_i)-\bm{b}\big\|_{\mathrm{RMSE}},
  \label{eq:reward}
\end{equation}

% \begin{equation}
%   r(y_i)
%   \;=\;
%   \nu(y_i)
%   \;-\;
%   \lambda
%   \sqrt{
%     \frac{1}{5}
%     \sum_{t \in \{O,C,E,A,N\}}
%     \left(
%       \hat b_t(y_i)-b_t
%     \right)^2
%   },
%   \label{eq:reward}
% \end{equation}

and form a preference pair $(y_w,y_l)$ from the candidates with the highest and lowest rewards. The trait term $-\lambda\,\mathrm{RMSE}(\hat{\bm{b}},\bm{b})$ is designed to discourage the alignment stage from washing out the target personality.

\paragraph{DPO objective.} We optimize $\theta$ with the Direct Preference
Optimization loss~\citep{rafailov2023dpo}, keeping $\phi$ and $\psi$ frozen:
\begin{equation}
  \mathcal{L}_\mathrm{DPO}(\theta)=
  -\,\mathbb{E}_{(\bm{b},y_w,y_l)}
  \log\sigma\!\Big(
     \beta\big[
       \Delta_\theta(y_w) - \Delta_\theta(y_l)
     \big]
  \Big),
\end{equation}
\begin{equation}
  \Delta_\theta(y)=
  \log\pi_\theta(y\mid\bm{b},x)-\log\pi_\mathrm{ref}(y\mid\bm{b},x),
\end{equation}
where $\sigma$ is the logistic function and $\beta$ the KL strength. Stage 2 keeps the personality encoder fixed and updates only the LoRA parameters, using a trait aware reward to encourage personality preservation.
% ======================================================================
%  Experiments  (requires: amsmath, booktabs)
% ======================================================================
\section{Experiments}
\label{sec:experiments}

% We design our experiments to answer the following research questions (RQs):
% \begin{itemize}
%   \item \textbf{RQ1 (Expression).} Does our method express a target Big Five
%         profile in natural dialogue, and how does it compare to
%         existing personality-control methods?
%   \item \textbf{RQ2 (Continuous control).} Is the induced trait continuously and
%         monotonically controllable, so that raising a target score raises the
%         measured trait?
%   \item \textbf{RQ3 (Generalization).} When we condition on a target profile,
%         does the model actually recover that profile when it takes a
%         questionnaire, including inventories and phrasings it did not see during
%         training?
%   \item \textbf{RQ4 (Ablation).} What does each component of our method (the
%         encoder, the survey loss, and the trait-anchored DPO) contribute?
% \end{itemize}

% Our experiments evaluate four aspects of FATE-5. We examine personality
% expression and dialogue naturalness, continuous trait control, recovery of
% target profiles under held out questionnaires, and the contribution of each
% model component.

\subsection{Data}
\label{sec:data}
The personality encoder and the trained baselines' conditioning modules are
learned from a corpus of examples, each providing a target profile $\bm{b}$, a
dialogue context $x$, and a response $y$ that matches $\bm{b}$. We build this
corpus from two complementary sources.

\textbf{Big5Chat}~\citep{big5chat} contributes persona grounded dialogues in which
a speaker is instructed to a high or low level of a single Big Five trait. We take
the utterance as the response $y$ and label it with the corresponding profile,
where the conditioned trait is set high or low and the others stay neutral. This
source gives clean and strongly expressed trait signals, but its targets are coarse
and categorical.

\textbf{PANDORA}~\citep{pandora} contributes real Reddit conversations. For each
comment written by a user whose Big Five profile is known, we recover its parent
comment as the context $x$ and use the comment itself as the response $y$,
labelled with the author's \emph{continuous} Big Five scores. This source provides fine-grained targets drawn from a real population, providing supervision across continuous trait values rather than only high or low settings.

In total the mixture contains about $46$k examples, roughly $30$k from Big5Chat and
$16$k from PANDORA, with all targets placed on a common $0$ to $100$ scale. We use
this mixture only to train the encoder, which is conditioned on $\bm{b}$ to produce
$y$ under the dialogue loss of \S\ref{sec:objective}. The survey loss instead draws
its items from the training questionnaires
$\mathcal{Q}_\mathrm{tr}=\{\text{BFI~\citep{bfi}},\text{IPIP-NEO-120~\citep{ipip-neo-120}}\}$. The
preference alignment stage of \S\ref{sec:dpo} uses no external corpus at all. Its
pairs come from on policy candidates that the encoder generates over random
profiles and dialogue topics, and an external judge scores them. For evaluation we
hold out two further questionnaires
$\mathcal{Q}_\mathrm{te}=\{\text{TIPI~\citep{tipi}},\text{BFI-2~\citep{bfi-2}}\}$, which appear in no training
stage.

\subsection{Models and Baselines}
\label{sec:models-baselines}

% \paragraph{Models.}
% We evaluate every method on three instruction-tuned backbones that span a scale
% ladder, \textbf{Llama-3-8B-Instruct}~\citep{llama3-8b}, \textbf{Mistral-Nemo-12B-Instruct}
% ~\citep{mistral-nemo-12b}, and \textbf{Mistral-Small-24B-Instruct}~\citep{mistral-small-24b}. Each method is trained and
% evaluated on each backbone independently, and the backbone weights are frozen for
% every method. The encoder uses $T{=}64$ soft-prompt tokens from a two-layer MLP with hidden size
% $512$ and is trained for $3\text{k}$ steps, and the offline DPO stage uses
% $\beta{=}0.1$, $\lambda{=}1$, and LoRA with $r{=}16$ and $\alpha{=}32$, on
% preference pairs built from $N{=}5$ candidates over $200$ random profiles.

\paragraph{Models.}
We evaluate every method on three instruction-tuned backbones from different
model families, \textbf{Llama-3-8B-Instruct}~\citep{llama3-8b},
\textbf{Gemma-2-9B-Instruct}~\citep{gemma2}, and
\textbf{OLMo-2-13B-Instruct}~\citep{olmo2}. Each method is trained and
evaluated on each backbone independently, and the backbone weights are frozen for
every method. The encoder uses $T{=}64$ soft-prompt tokens from a two-layer MLP with hidden size
$512$ and is trained for $3\text{k}$ steps, and the offline DPO stage uses
$\beta{=}0.1$, $\lambda{=}1$, and LoRA with $r{=}16$ and $\alpha{=}32$, on
preference pairs built from $N{=}5$ candidates over $200$ random profiles. The candidates are scored for the DPO reward by Llama-3.1-70B-Instruct~\citep{llama3-8b}, which is distinct from the judges used for evaluation (\S\ref{sec:eval}).

\paragraph{Baselines.}
We compare against four representative families of controllable personality
generation, all conditioned on the same target $\bm{b}\in[0,100]^5$:
\begin{itemize}
  \item \textbf{Prompting (Big5-scaler)~\citep{cho2025scaling}.} The target scores are verbalized in the
        system prompt and no parameters are trained.
  \item \textbf{Steering vectors~\citep{steering-vector}.} Per-trait directions extracted from the
        backbone are added to the residual stream of the middle layers, scaled by
        the target profile.
  \item \textbf{PsychAdapter~\citep{psychadapter}.} A trained module maps $\bm{b}$ to per-layer
        additive vectors injected at the middle layers, with the backbone frozen.%수정필요
  \item \textbf{DPG~\citep{dpg}.} A hypernetwork maps $\bm{b}$ to a low-rank weight modulation
        (LoRA) applied to the attention output projections, with the backbone
        frozen.
\end{itemize}
The trained baselines use the same personality dialogue corpus as ours but do not use the survey loss. Appendix~\ref{app:setup} gives the full training and decoding configuration.

\subsection{Evaluation}
\label{sec:eval}
We assess each method along two complementary axes, how the target personality is
\emph{expressed} in free dialogue and how faithfully it is \emph{measured} by
standardized questionnaires. Also, we ablate our own design. Each part answers one of
the research questions.

\subsubsection{Dialogue expression (RQ1).}
Two agents converse over a shared topic for $10$ turns, with each agent assigned
an independent random target profile. We use ten conversation topics, two for
each Big Five trait, chosen to naturally elicit the trait of interest. Openness
is probed by travel and music, Conscientiousness by daily habits and personal
goals, Extraversion by making friends and social events, Agreeableness by a
disagreement with a friend and helping a stranger in need, and Neuroticism by
facing an unexpected setback and dealing with a stressful deadline. Each dialogue
is evaluated only on the trait targeted by its topic. For each method and
backbone, we generate $100$ dialogues balanced across the ten topics.

We evaluate dialogue expression using two metrics:
\begin{itemize}
  \item \textbf{Trait accuracy.} Given a dialogue, the evaluator selects which
        of the two agents expresses more of the target trait. Accuracy is the
        fraction of responses that match the ground-truth ordering of the two
        target scores.
  \item \textbf{Naturalness.} The evaluator rates how human-like and fluent the
        dialogue is on a $1$--$5$ scale, which we rescale to $0$--$100$.
\end{itemize}

For the main evaluation, we use three external LLM judges
(GPT-5-mini, Claude-Haiku-4.5, and DeepSeek-V4-Flash) and average their scores.
We report the results on all three backbones.

We additionally conduct a human evaluation on $50$ Llama-3-8B dialogues using
the same persona and topic configurations across methods. We collect $984$
dialogue level evaluations from $59$ unique workers. Human raters evaluate Trait
and Naturalness using the same criteria as the LLM judges, with an additional
``I don't know'' option for Trait. These responses account for $5.2\%$ of Trait
judgments and are excluded from Trait accuracy. Aggregated ratings show high
split half reliability, reaching $0.90$ for Trait and $0.94$ for Naturalness.
Further study details are provided in Appendix~\ref{app:human}.

\subsubsection{Questionnaire measurement (RQ2 and RQ3).}
To measure how faithfully each method encodes the target, and whether this
survives a change of instrument, we administer the held-out inventories
$\mathcal{Q}_\mathrm{te}$. Given a target $\bm{b}$, we condition the method, ask
each item in a natural conversational phrasing, parse the rating, and apply the
questionnaire's rubric to recover an estimated profile $\hat{\bm{b}}$. We report:
\begin{itemize}
  \item \textbf{Monotonicity ($r$), RQ2.} Sweeping one trait's target from $0$ to
        $100$ (others fixed at $50$), the Pearson correlation between the target
        and the recovered score for that trait, capturing \emph{directional
        control}.
  \item \textbf{Recovery error (RMSE), RQ3.} The RMSE between $\hat{\bm{b}}$ and
        $\bm{b}$ ($0$--$100$) over $25$ random profiles, capturing absolute
        agreement or \emph{calibration}.
\end{itemize}
Because self-report from LLMs is known to be phrasing-sensitive, We evaluate monotonicity under both a rigid “output a single digit” format and a natural conversational format, and report recovery error under the natural format. These experiments are conducted on
Llama-3-8B, Gemma-2-9B, and OLMo-2-13B.

\subsubsection{Ablation of our method (RQ4).}
On Llama-3-8B we remove or substitute one component at a time and re-score with the
dialogue metrics, to attribute the roles of the encoder, the survey loss, and the
DPO stage:
\begin{itemize}
  \item \textbf{Encoder $\times$ DPO ($2{\times}2$).} Our full model
        (encoder\,+\,DPO) versus the encoder alone ($-$DPO), a numeric text prompt
        with DPO ($-$encoder), and plain prompting ($-$encoder, $-$DPO), comparing learned soft-prompt conditioning against numeric prompting under matched DPO and non-DPO settings.
  \item \textbf{$-$ Survey loss.} The encoder is retrained without
        $\mathcal{L}_\mathrm{surv}$ (then aligned with DPO), testing whether the
        survey loss is what makes the trait faithfully encoded.
  \item \textbf{Reward trait term.} Setting $\lambda{=}0$ removes the trait anchor
        from the DPO reward (Eq.~\ref{eq:reward}); we additionally sweep
        $\lambda{=}2$ to trace the trait and naturalness trade-off.
  \item \textbf{DPO $\rightarrow$ SFT.} Replacing DPO with supervised fine-tuning
        on the preferred candidate, testing whether the contrastive preference
        signal matters beyond imitation.
\end{itemize}

We provide examples of the numeric prompts used in the ablation in Appendix~\ref{app:numeric}.

% ======================================================================
%  Results  (requires: amsmath, booktabs)
% ======================================================================
\section{Results}
\label{sec:results}

% \begin{table*}[t]
% \centering
% \small
% \setlength{\tabcolsep}{4.5pt}
% \begin{adjustbox}{max width=\textwidth}
% \begin{tabular}{ll ccc ccc ccc ccc}
% \toprule
% & & \multicolumn{9}{c}{\textbf{LLM judge}} & \multicolumn{3}{c}{\textbf{Human}} \\
% \cmidrule(lr){3-11}\cmidrule(lr){12-14}
% & & \multicolumn{3}{c}{\textbf{Llama-3-8B}}
% & \multicolumn{3}{c}{\textbf{Gemma-2-9B}}
% & \multicolumn{3}{c}{\textbf{OLMo-2-13B}}
% & \multicolumn{3}{c}{\textbf{Llama-3-8B}} \\
% \cmidrule(lr){3-5}\cmidrule(lr){6-8}\cmidrule(lr){9-11}\cmidrule(lr){12-14}
% & Method
% & Trait & Nat. & Avg.
% & Trait & Nat. & Avg.
% & Trait & Nat. & Avg.
% & Trait & Nat. & Avg. \\
% \midrule
% \multirow{4}{*}{\rotatebox{90}{baseline}}
% & Prompting (big5-scaler)
% & 63.7 & \underline{39.4} & \underline{51.5}
% & 65.3 & \textbf{54.1} & \textbf{59.7}
% & 51.3 & \textbf{39.8} & 45.6
% & 73.3 & \textbf{74.4} & \textbf{73.8} \\

% & Steering vector
% & \textbf{77.3} & 2.6 & 39.9
% & \underline{74.3} & 7.9 & 41.1
% & \textbf{80.3} & 21.6 & \underline{51.0}
% & \underline{78.5} & 20.2 & 49.3 \\

% & PsychAdapter
% & \underline{77.0} & 14.7 & 45.9
% & \textbf{78.7} & 28.2 & 53.4
% & 69.7 & 21.0 & 45.3
% & 71.2 & 53.6 & 62.4 \\

% & DPG
% & 66.7 & 33.8 & 50.2
% & 63.0 & 34.8 & 48.9
% & 69.7 & 26.9 & 48.3
% & 69.9 & 66.2 & 68.0 \\
% \midrule

% & \textbf{FATE-5}
% & 75.3 & \textbf{43.0} & \textbf{59.1}
% & 70.0 & \underline{48.4} & \underline{59.2}
% & \underline{72.0} & \underline{38.9} & \textbf{55.5}
% & \textbf{78.7} & \underline{67.8} & \underline{73.3} \\

% \bottomrule
% \end{tabular}
% \end{adjustbox}

\begin{table*}[t]
\centering
\small
\setlength{\tabcolsep}{4.5pt}
\begin{adjustbox}{max width=\textwidth}
\begin{tabular}{ll cc cc cc cc}
\toprule
& & \multicolumn{6}{c}{\textbf{LLM judge}} & \multicolumn{2}{c}{\textbf{Human}} \\
\cmidrule(lr){3-8}\cmidrule(lr){9-10}
& & \multicolumn{2}{c}{\textbf{Llama-3-8B}}
& \multicolumn{2}{c}{\textbf{Gemma-2-9B}}
& \multicolumn{2}{c}{\textbf{OLMo-2-13B}}
& \multicolumn{2}{c}{\textbf{Llama-3-8B}} \\
\cmidrule(lr){3-4}\cmidrule(lr){5-6}\cmidrule(lr){7-8}\cmidrule(lr){9-10}
& Method
& Trait Acc. & Nat.
& Trait Acc. & Nat.
& Trait Acc. & Nat.
& Trait Acc. & Nat. \\
\midrule
\multirow{4}{*}{\rotatebox{90}{baseline}}
& Prompting (big5-scaler)
& 63.7 & \underline{39.4}
& 65.3 & \textbf{54.1}
& 51.3 & \underline{39.8}
& 73.3 & \textbf{74.4} \\

& Steering vector
& \textbf{77.3} & 2.6
& 74.3 & 7.9
& \textbf{80.3} & 21.6
& \underline{78.5} & 20.2 \\

& PsychAdapter
& \underline{77.0} & 14.7
& \textbf{78.7} & 28.2
& 69.7 & 21.0
& 71.2 & 53.6 \\

& DPG
& 66.7 & 33.8
& 63.0 & 34.8
& 69.7 & 26.9
& 69.9 & 66.2 \\
\midrule

& \textbf{FATE-5}
& 75.3 & \textbf{43.0}
& \underline{76.3} & \underline{39.1}
& \underline{72.7} & \textbf{45.4}
& \textbf{78.7} & \underline{67.8} \\

\bottomrule
\end{tabular}
\end{adjustbox}

\caption{
Dialogue expression across three backbones, evaluated by LLM judges and humans.
Trait Acc. measures whether the evaluator correctly identifies which agent has the higher target-trait level, and Nat. measures dialogue naturalness on a $0$--$100$ scale.
Best results are bolded and second-best results are underlined.
}
\label{tab:main}
\end{table*}

\subsection{RQ1: Expression}
\label{sec:rq1}

Table~\ref{tab:main} reports dialogue expression across the three backbones.
Activation injection methods achieve high trait accuracy but substantially
reduce naturalness. For example, Steering reaches high trait accuracy while its
naturalness remains low at $2.6$, $7.9$, and $21.6$ across Llama-3-8B,
Gemma-2-9B, and OLMo-2-13B. PsychAdapter improves naturalness compared with
Steering but still remains below our method across all three backbones.
Prompting generally provides lower trait accuracy and shows mixed naturalness
across backbones.

Our method maintains high trait accuracy, ranging from $72.7$ to $76.3$ across
the three backbones. Its naturalness scores are $43.0$, $39.1$, and $45.4$ on
Llama-3-8B, Gemma-2-9B, and OLMo-2-13B, respectively. These results indicate
that our method preserves perceptible differences in target-trait levels while
avoiding much of the naturalness degradation observed with activation-based
approaches.

The human evaluation on Llama-3-8B shows a similar pattern. Steering achieves
high trait accuracy but substantially lower naturalness, while prompting
achieves the highest naturalness with lower trait accuracy. Our method obtains
the highest trait accuracy at $78.7$ while achieving a naturalness score of
$67.8$, compared with $74.4$ for prompting. The human evaluation therefore
shows that our method maintains high trait accuracy while preserving relatively
high dialogue naturalness.

% \subsection{RQ2: Continuous control}
% \label{sec:rq2}

% Table~\ref{tab:survey} reports the monotonicity of each method, measured
% as the Pearson correlation between the target trait value and the score recovered
% from held-out questionnaires. FATE-5 achieves a correlation of $0.99$, and
% this result remains stable across different question formats. The baselines show
% a much stronger dependence on phrasing. Under natural conversational questions,
% PsychAdapter, DPG, and Steering obtain correlations of $0.67$, $0.74$, and
% $0.45$, respectively. Under a rigid single-digit response format, these values
% drop to $-0.10$, $0.20$, and $0.15$.

% FATE-5 maintains strong directional control under both formats. Prompting
% also preserves this relationship to a lesser extent, while the injection-based
% baselines become much less reliable when the response format is constrained.

\subsection{RQ2: Continuous Control}
\label{sec:rq2}

Table~\ref{tab:survey} reports monotonicity across all three backbones, measured
as the Pearson correlation between the target trait value and the score recovered
from held out questionnaires. FATE-5 shows consistently strong directional
control under both question formats. It reaches $0.99$ under both rigid and
natural phrasing on Llama-3-8B and Gemma-2-9B, and $0.97$ and $0.98$ on
OLMo-2-13B.

The baselines are less consistent across backbones and response formats.
Prompting shows strong monotonicity in several settings but drops to $0.65$
under natural phrasing on OLMo-2-13B. The injection based methods also vary
substantially across models and formats. Overall, FATE-5 shows consistently high target correlations across the evaluated backbones and questionnaire formats.

\begin{table*}[t]
\centering
\small
\begin{adjustbox}{max width=\textwidth}
\begin{tabular}{l ccc ccc ccc}
\toprule
& \multicolumn{3}{c}{\textbf{Llama-3-8B}}
& \multicolumn{3}{c}{\textbf{Gemma-2-9B}}
& \multicolumn{3}{c}{\textbf{OLMo-2-13B}} \\
\cmidrule(lr){2-4}\cmidrule(lr){5-7}\cmidrule(lr){8-10}
& \multicolumn{2}{c}{Monot. $r\uparrow$} & Recov.
& \multicolumn{2}{c}{Monot. $r\uparrow$} & Recov.
& \multicolumn{2}{c}{Monot. $r\uparrow$} & Recov. \\
\cmidrule(lr){2-3}\cmidrule(lr){5-6}\cmidrule(lr){8-9}
Method & rigid & natural & RMSE $\downarrow$
       & rigid & natural & RMSE $\downarrow$
       & rigid & natural & RMSE $\downarrow$ \\
\midrule
Prompting (big5-scaler) & 0.78 & 0.93 & 20.7 & 0.92 & 0.86 & 18.0 & 0.94 & 0.65 & 25.4 \\
Steering vector         & 0.15 & 0.45 & 28.8 & 0.68 & 0.74 & 28.6 & 0.39 & 0.78 & 26.8 \\
PsychAdapter            & $-0.10$ & 0.67 & 28.8 & 0.45 & 0.36 & 39.8 & 0.76 & 0.66 & 27.3 \\
DPG                     & 0.20 & 0.74 & 28.5 & 0.45 & 0.48 & 30.8 & 0.43 & 0.69 & 28.7 \\
\textbf{FATE-5}         & \textbf{0.99} & \textbf{0.99} & \textbf{10.5} & \textbf{0.99} & \textbf{0.99} & \textbf{8.9} & \textbf{0.97} & \textbf{0.98} & \textbf{12.4} \\
\bottomrule
\end{tabular}
\end{adjustbox}
\caption{Questionnaire measurement on all three backbones (RQ2, RQ3), using
held-out inventories TIPI and BFI-2. Monotonicity is the Pearson correlation
between target and recovered trait, under a rigid and a natural phrasing.
Recovery RMSE ($0$--$100$) is the absolute error between the recovered and target
profiles under the natural phrasing. For reference, always predicting the midpoint (50) yields an expected RMSE of 28.9 for uniformly sampled targets.}
\label{tab:survey}
\end{table*}

% \subsection{RQ3: Questionnaire Generalization}
% \label{sec:rq3}

% Directional control is necessary but not sufficient. Table~\ref{tab:survey}
% reports the recovery error by measuring the absolute agreement between the
% recovered and target profiles. The baselines differ substantially from FATE-5
% on this metric. A method that ignores the target reaches an RMSE of about $28.9$.
% The injection based baselines remain close to this level, with PsychAdapter at
% $28.8$, Steering at $28.8$, and DPG at $28.5$, even when natural phrasing yields
% positive monotonicity.

% These methods can move the recovered trait in the correct direction, but they do
% not accurately reach the intended target value. Their recovered scores remain
% compressed within a relatively narrow range. Prompting achieves a lower recovery
% error of $20.7$, while FATE-5 further reduces it to $10.5$. Taken together, the
% results for RQ2 and RQ3 show that FATE-5 not only follows the target direction
% but also recovers values close to the intended profile across different question
% formats.

\subsection{RQ3: Questionnaire Generalization}
\label{sec:rq3}

Directional control alone does not guarantee that the recovered personality
matches the intended target value. Table~\ref{tab:survey} therefore also reports
recovery RMSE across all three backbones. FATE-5 achieves the lowest error in
every setting, with RMSE values of $10.5$, $8.9$, and $12.4$ on Llama-3-8B,
Gemma-2-9B, and OLMo-2-13B, respectively.

The injection based baselines generally remain close to or above the
no-control reference of about $28.9$ RMSE, while prompting performs better with
errors of $20.7$, $18.0$, and $25.4$. FATE-5 substantially reduces these errors
across all three backbones. Taken together, the results for RQ2 and RQ3 show
that FATE-5 not only follows the intended trait direction but also recovers
personality profiles substantially closer to the target values across models
.

\subsection{RQ4: Ablation}
\label{sec:ablation}

Table~\ref{tab:ablation} reports the ablation results for FATE-5 on Llama-3-8B. The two-by-two comparison of the encoder and DPO stage highlights their distinct roles. Removing DPO while retaining the encoder lowers naturalness from $43.0$ to $37.5$, a decrease of $5.5$ points. This suggests that DPO contributes to improving dialogue naturalness. Removing the encoder while applying the same DPO procedure to a numeric prompt reduces trait accuracy from $75.3$ to $60.0$, a decrease of $15.3$ points. This shows that the encoder plays a key role in effective trait expression. At the same time, this configuration achieves the highest naturalness score of $48.8$. The plain numeric prompt without either component reaches a trait accuracy of $69.0$ and a naturalness score of $46.2$. These results suggest that the encoder strengthens trait expression at the cost of some dialogue naturalness. This motivates the subsequent DPO stage, which compensates for this loss in naturalness while preserving the trait expression learned by the encoder. These results indicate that the learned personality representation and preference alignment work jointly: DPO improves naturalness when applied to the encoder representation while maintaining, and in this setting slightly improving, trait expression.

Removing the survey loss reduces trait accuracy from $75.3$ to $51.3$, supporting its importance for faithful personality encoding. The trait component of the reward is also important. Setting $\lambda=0$ lowers both trait accuracy and naturalness. As detailed in Appendix~\ref{app:lambda0}, removing the continuous trait term makes the reward highly discrete, producing many ties and reducing the number of usable preference pairs from 198 to 37. Thus, the trait term not only anchors personality agreement but also provides a denser preference signal. Increasing the weight to $\lambda=2$ raises trait accuracy to $78.7$ but reduces naturalness to $32.5$, illustrating the tradeoff induced by the reward weight. Finally, replacing DPO with SFT reduces both trait accuracy and naturalness, suggesting that the contrastive preference objective provides additional benefit.

\begin{table}[t]
\centering
\small
\begin{tabular}{l cc}
\toprule
Configuration & Trait Acc. & Nat. \\
\midrule
\textbf{FATE-5 (Encoder+DPO, $\lambda{=}1$)} & 75.3 & 43.0 \\
\midrule
$-$ DPO (Encoder only)        & 71.3 & 37.5 \\
$-$ Encoder (numeric + DPO)   & 60.0 & 48.8 \\
$-$ Enc $-$ DPO (numeric)     & 69.0 & 46.2 \\
$-$ Survey loss               & 51.3 & 34.2 \\
DPO $\rightarrow$ SFT         & 69.7 & 41.3 \\
\midrule
Reward $\lambda{=}0$          & 62.7 & 19.4 \\
Reward $\lambda{=}2$          & 78.7 & 32.5 \\
\bottomrule
\end{tabular}
\caption{Ablation of FATE-5 on Llama-3-8B (RQ4), scored using the same LLM judges as Table~\ref{tab:main}, which are distinct from the judge used to build the DPO preference pairs. Trait Acc. measures target-trait accuracy, and Nat. measures dialogue naturalness on a $0$--$100$ scale. The full model row is taken from the Llama-3-8B column of Table~\ref{tab:main}. Each other row removes or substitutes one component.}
\label{tab:ablation}
\end{table}
\section{Conclusion}

We introduced FATE-5, a framework for faithful and natural continuous Big Five
personality control that separates personality representation from naturalness
alignment. Experiments across three backbones and human evaluation show that
FATE-5 maintains strong trait expression while producing substantially more
natural dialogue than other trait-strong approaches. 
Psychometric evaluation
further demonstrates reliable continuous control and substantially lower profile
recovery error, while ablations support the contributions of survey supervision
and preference alignment. 
These findings suggest that personality fidelity and natural dialogue quality, which are difficult to achieve simultaneously with existing methods, can be jointly maintained in personality-controllable conversational LLMs.
%We introduced FATE-5, a framework for faithful and natural continuous Big Five personality control. By separating personality representation from naturalness alignment, FATE-5 offers a favorable balance between trait expression and dialogue naturalness. Experiments across three backbones and human evaluation show that FATE-5 maintains strong trait expression while producing substantially more natural dialogue than other trait-strong approaches. Psychometric evaluation further demonstrates reliable continuous control and substantially lower profile recovery error, while ablations verify the complementary roles of survey supervision and trait-preserving naturalness alignment. 
%By simultaneously preserving personality fidelity and natural dialogue quality, two goals that are difficult to achieve together with existing methods, FATE-5 offers a practical path toward personality-controllable conversational LLMs.

\section*{AI Use Statement}

In this work, we used generative AI tools to assist with parts of the experimental process, including code development and debugging, and to support manuscript writing and language refinement. All AI-assisted experimental procedures, code, analyses, and written content were reviewed and verified by the authors. Generative AI was not used as a substitute for scientific judgment or independent validation of the results. The authors take full responsibility for the final content of the paper, including all experimental results, claims, and AI-assisted text.

\section*{Ethics Statement}

The human evaluation conducted in this study was reviewed and approved by the Institutional Review Board prior to data collection. All participants provided informed consent before taking part in the study, and participation was voluntary. The evaluation was conducted through Amazon Mechanical Turk, and participant responses were used only for the purposes of this research. No personally identifiable information was included in the analysis or reported in this paper.

\subsection*{Reproducibility statement}

We provide the details needed to reproduce our experiments throughout the
paper and appendix. The model architecture and training objectives are
described in Section~3, while the datasets, baselines, and evaluation protocols
are described in Section~4. Appendix~A provides details of the survey
supervision procedure, and Appendix~B reports the full training, preference
construction, baseline implementation, and generation settings. All experiments
use fixed random seeds, and the model and optimization configurations are
reported for each training stage.

% \subsubsection*{Author Contributions}
% If you'd like to, you may include  a section for author contributions as is done
% in many journals. This is optional and at the discretion of the authors.

% \subsubsection*{Acknowledgments}
% Use unnumbered third level headings for the acknowledgments. All
% acknowledgments, including those to funding agencies, go at the end of the paper.

\bibliography{iclr2027_conference}
\bibliographystyle{iclr2027_conference}

\appendix
% \section{Appendix}

\section{Pseudo-Answer Construction for the Survey Loss}
\label{app:pseudo}

The survey loss (Eq.~\ref{eq:surv}) requires, for a target profile $\bm{b}$ and a
questionnaire item $q_j$, an answer $\tilde{a}_j$ that a respondent with profile
$\bm{b}$ would give. Since no human responses exist for arbitrary continuous
targets, we derive $\tilde{a}_j$ deterministically by inverting the
questionnaire's scoring rubric $\rho$.

\paragraph{Target sampling.}
Survey targets are sampled independently of the dialogue corpus. We draw $200$
random profiles with each trait uniform on the $1$--$5$ scale,
$t_k \sim \mathcal{U}(1,5)$, and pass the equivalent $0$--$100$ profile
$b_k = 100\,(t_k-1)/4$ to the encoder.

\paragraph{Item-level answers.}
Let $k(j)$ be the trait that item $q_j$ measures and
$p_j = (t_{k(j)}-1)/4 \in [0,1]$ the normalized target of that trait.
For a questionnaire with answer options $\{1,\dots,M\}$, the pseudo-answer is the
nearest option on a linear map of $p_j$,
\begin{equation}
  \tilde{a}_j =
  \begin{cases}
    \mathrm{round}\big(1 + (M-1)\,p_j\big), & q_j \text{ positively keyed},\\
    \mathrm{round}\big(M - (M-1)\,p_j\big), & q_j \text{ reverse keyed},
  \end{cases}
\end{equation}
clipped to $[1,M]$. Thus a high target yields strong agreement on positively
keyed items and strong disagreement on reverse-keyed items, and a mid-range
target yields the neutral option. Because every item of a trait receives the
answer implied by the same $t_{k(j)}$, applying the rubric recovers the target up
to rounding, $\rho(\tilde{a}) \approx \bm{b}$. We use $M{=}5$ for BFI and
IPIP-NEO-120. Items are assigned to traits and keyed
according to each inventory's published scoring key.

\paragraph{Training procedure.}
At each training step, one questionnaire is selected from
$\mathcal{Q}_{\mathrm{tr}}$ in rotation, and $8$ of its items are sampled for
each target in the batch. Each item is presented as a single-item prompt,
\begin{quote}\small
\texttt{Statement: "<item>"}\\
\texttt{Rate from 1 (strongly disagree) to $M$ (strongly agree). Output only the single digit.}
\end{quote}
with the soft prompt $E_\phi(\bm{b})$ prepended, and $\ell$ in
Eq.~\ref{eq:surv} is the negative log-likelihood of the answer digit
$\tilde{a}_j$ only; the prompt tokens are masked from the loss. The survey loss
is enabled after a warm-up of $300$ steps trained on the dialogue loss alone,
after which the two losses are combined with $\lambda_{\mathrm{dial}}{=}0.6$ and
$\lambda_{\mathrm{surv}}{=}0.4$ (Eq.~\ref{eq:full}).

\paragraph{Separation from evaluation.}
Pseudo-answers are built only from the training inventories
$\mathcal{Q}_{\mathrm{tr}}$ and the rigid single-digit format above.
The evaluation inventories $\mathcal{Q}_{\mathrm{te}}$ (TIPI, BFI-2) and the
conversational phrasing used in §\ref{sec:eval} never appear in training, so the
questionnaire results in Table~\ref{tab:survey} measure transfer to unseen items
and formats rather than memorized answers.

\section{Training and Reproduction Details}
\label{app:setup}

All methods are trained and evaluated separately for each backbone while the
backbone weights remain frozen. All runs use bfloat16 on a single GPU with seed
$42$.

\paragraph{Personality encoder.}
Stage 1 uses a two layer MLP with GELU activations and hidden size $512$. The
encoder produces $T{=}64$ soft prompt tokens and is trained for $3\text{k}$
steps. We use AdamW with a learning rate of $10^{-3}$ and weight decay of
$0.01$. The learning rate follows a linear schedule with $100$ warmup steps.
The batch size is $2$ and gradient accumulation is $4$, giving an effective
batch size of $8$. The maximum sequence length is $512$ tokens.

The first $300$ steps use only the dialogue loss as a warmup. After this period,
we combine the two losses using $\lambda_{\mathrm{dial}}{=}0.6$ and
$\lambda_{\mathrm{surv}}{=}0.4$. At each survey step, we rotate through
$\mathcal{Q}_{\mathrm{tr}}$ and sample $8$ items for each target. The targets
come from $200$ random profiles sampled uniformly over the answer scale.
Pseudo answers are constructed deterministically as described in
Appendix~\ref{app:pseudo}.

\paragraph{Preference alignment.}
Stage 2 constructs preference pairs from the frozen encoder using $200$ random
profiles paired with the ten dialogue topics. For each profile, we sample
$N{=}5$ candidates. Generation uses temperature $1.0$, top $p$ of $0.95$,
repetition penalty $1.15$, no repeat $4$ grams, and at most $120$ new tokens.
We score the candidates using Eq.~\ref{eq:reward} with $\lambda{=}1$. The
candidate with the highest reward is selected as $y_w$ and the candidate with
the lowest reward is selected as $y_l$. Pairs with a reward gap of zero are
discarded. We use Llama-3.1-70B-Instruct as the scoring judge.

DPO trains a LoRA adapter with rank $16$, scaling parameter $\alpha{=}32$, and
dropout $0.05$. The adapter is applied to all attention and MLP projections.
Training runs for $200$ optimizer steps with AdamW and a learning rate of
$5\times10^{-5}$. We use $\beta{=}0.1$, gradient clipping at $1.0$, and a
maximum sequence length of $1024$ tokens. For Llama-3-8B, the batch size is $2$
and gradient accumulation is $4$. For the larger backbones, the batch size is
$1$ and gradient accumulation is $8$. Both settings give an effective batch
size of $8$. The encoder and backbone remain frozen throughout this stage.

\paragraph{Baselines.}
No official implementation directly supports our setting of continuous Big Five
conditioning with a frozen instruction tuned backbone. We therefore reimplement
the three trained baselines following their original papers and train them on
the same corpus for the same $3\text{k}$ steps. We use the same optimizer,
effective batch size, and sequence length as for our encoder. For consistency, each uses a two layer MLP
with hidden size $512$ that maps $\bm{b}$ to the injected parameters. This
design allows the comparison to focus on the conditioning mechanism rather than
model size or training budget.

\begin{itemize}
  \item \textbf{PsychAdapter.}
        The MLP maps $\bm{b}$ to one additive vector for each layer in the
        middle $25$ to $75\%$ of the network. The vectors pass through
        $\tanh$ and are scaled by $8$ to bound the perturbation. The learning
        rate is $5\times10^{-4}$.

  \item \textbf{DPG.}
        A hypernetwork maps $\bm{b}$ to a rank $8$ LoRA modulation of the
        attention output projection in every layer. The modulation is applied
        with scale $0.5$. The learning rate is $2\times10^{-4}$.

  \item \textbf{Steering vectors.}
        Directions are extracted from $300$ profiles. Each profile is prompted
        with a numeric persona and one of generated open ended questions. For every
        layer and trait, we compute the covariance direction between the trait
        score and the hidden state of the final token and normalize the
        direction to unit norm. During generation, these directions are added to the residual stream in the middle $25$ to $75\%$ of layers. For trait $t$, the target score $b_t$ is
mapped from $[0,100]$ to $[-1,1]$, and the steering coefficient is
$\alpha_t = c(2b_t/100 - 1)$, where $c$ controls the overall injection strength.

  \item \textbf{Prompting with Big5 Scaler.}
        We use the prompt format published by the authors without additional
        training.
\end{itemize}

For each baseline, we select the learning rate and injection scale from a small
grid using a held out subset of the training corpus. We choose the setting with
the lowest validation loss among those that do not produce degenerate text. The
values reported above are the selected settings.

\paragraph{Dialogue generation.}
All methods use the same decoding settings. We use temperature $0.7$, top $p$
of $0.9$, repetition penalty $1.15$, no repeat $4$ grams, and at most $120$ new
tokens per turn. Each dialogue contains $10$ turns and uses a context window
consisting of the most recent $6$ utterances. We regenerate a dialogue if it
ends before $8$ utterances. If generation still fails after $5$ attempts, we
resample the persona and topic pair.

\section{Human Evaluation Setting}
\label{app:human}

\paragraph{Sampling.} From the $100$ dialogues generated per method, we select $10$ dialogues using a fixed random seed, with one dialogue selected for each topic. The same ten persona and topic configurations are used for all five methods. As a result, the $50$ dialogues differ only in the generation method.

\paragraph{Instrument.} The $50$ dialogues are split across three forms containing $17$, $17$, and $16$ dialogues. Each form includes dialogues from every method. Dialogue order is shuffled, method identities are hidden, and the agents are presented only as Speaker A and Speaker B. Each dialogue is rated using the same two criteria as the LLM judges. Trait is evaluated through a forced choice question with an additional ``I don't know'' option. Naturalness is rated on a $1$ to $5$ scale and is rescaled to $0$ to $100$ using the same rescaling as the LLM judges. Each item presents only a behavioral description of the target trait, such as ``outgoing, talkative and energetic,'' and does not include the trait label.

\paragraph{Participants.} We recruit workers through Amazon Mechanical Turk and restrict participation to workers with a lifetime HIT approval rate above $98\%$. Workers are paid \$3.00 per form. Each worker completes exactly one form, so no worker evaluates the same dialogue more than once.

\paragraph{Exclusions.} We use $59$ submissions from $59$ unique workers and $984$ dialogue level evaluations. At the response level, ``I don't know'' answers account for $5.2\%$ of Trait judgments. These responses are excluded from Trait accuracy but are included in the reported response totals.

\paragraph{Reliability.} Inter rater agreement is $\alpha{=}0.49$ for Trait and $\alpha{=}0.45$ for Naturalness. We use the nominal form of Krippendorff's alpha for Trait and the interval form for Naturalness. Because each dialogue is rated by approximately twenty workers, the aggregated dialogue level scores are considerably more stable than individual ratings. Split half reliability with the Spearman--Brown correction reaches $0.94$ for Naturalness and $0.90$ for Trait. Human and LLM judge naturalness scores also show strong agreement at the level of individual dialogues. Across the $50$ dialogues, the Pearson correlation is $r{=}0.84$ and the Spearman correlation is $\rho{=}0.85$. This agreement supports the use of LLM judges for the backbones that were not evaluated with human raters. We report $95\%$ confidence intervals using $10{,}000$ cluster bootstrap resamples at the dialogue level.

\section{Numeric prompt used in the ablation}
\label{app:numeric}

\begin{center}
\fbox{\begin{minipage}{0.93\columnwidth}
\small\ttfamily
You are a person with the following Big5 personality profile:\\[2pt]
\hspace*{2em}Openness: 75/100 (high) --- imaginative, curious, and open to\\
\hspace*{6.2em}new experiences\\
\hspace*{2em}Conscientiousness: 20/100 (very low) --- spontaneous, flexible,\\
\hspace*{6.2em}and less structured\\
\hspace*{2em}Extraversion: 60/100 (moderate) --- outgoing, energetic, and\\
\hspace*{6.2em}socially enthusiastic\\
\hspace*{2em}Agreeableness: 45/100 (moderate) --- skeptical, competitive, and\\
\hspace*{6.2em}less inclined to compromise\\
\hspace*{2em}Neuroticism: 30/100 (low) --- emotionally stable, calm, and\\
\hspace*{6.2em}resilient under pressure\\[2pt]
Express these personality traits naturally in conversation.
\end{minipage}}
\end{center}

\section{Why Removing the Trait Anchor Also Hurts Naturalness}
\label{app:lambda0}

Setting $\lambda{=}0$ leaves only the naturalness score from the judge in the
reward. One might therefore expect naturalness to improve. Instead, it drops
sharply as shown in Table~\ref{tab:ablation}. This behavior originates from the
preference data rather than from the optimization process.

The judge evaluates naturalness on a scale from $1$ to $5$. When
$\lambda{=}0$, the reward can therefore take only five values and every reward
gap is a multiple of $25$. If all $N{=}5$ candidates for a profile receive the
same score, the reward gap becomes zero and the pair is discarded. This occurs
for most profiles. Among the $200$ sampled profiles, only $37$ profiles produce
a usable pair, which corresponds to $18.5\%$ of the data. In contrast,
$\lambda{=}1$ produces usable pairs for $198$ profiles, which corresponds to
$99\%$ of the data. Table~\ref{tab:pairstats} summarizes these statistics. The
trait term provides a finer-grained signal that breaks these ties and allows
almost every profile to contribute to preference training.

\begin{table}[t]
\centering
\small
\begin{tabular}{l ccc cc}
\toprule
$\lambda$ & Usable pairs & Mean reward gap & Mean utterance length
& Nat.\\
\midrule
$0$   & $37$  & $33.8$ & $12.3$ & $19.4$ \\
$1$   & $198$ & $12.8$ & $36.3$ & $43.0$ \\
$2$   & $200$ & $25.7$ & $34.4$    & $32.5$ \\
\bottomrule
\end{tabular}
\caption{Preference pair statistics for different values of $\lambda$ on
Llama-3-8B. With $\lambda{=}0$, the reward is limited to five discrete levels.
This produces many ties between candidates and removes most profiles from
preference training, resulting in a much smaller preference-training set. Mean reward gap is reported only as a within-setting diagnostic of pair separation and is not directly comparable across different values of \(\lambda\), since the reward definition changes with \(\lambda\). Mean utterance length is measured in tokens.}
\label{tab:pairstats}
\end{table}

\end{document}